\documentclass{article}

\usepackage{microtype}
\usepackage{graphicx}
\usepackage{subfigure}
\usepackage{booktabs} 

\usepackage{hyperref}

\usepackage[accepted]{icml2023}

\usepackage{amsmath}
\usepackage{amssymb}
\usepackage{mathtools}
\usepackage{amsthm}

\usepackage{microtype}
\usepackage{CJKutf8} 
\usepackage{graphicx}
\usepackage{subfigure}
\usepackage{booktabs} 
\usepackage{graphicx}
\usepackage{multirow}

\usepackage{hyperref}
\usepackage{enumitem}

\usepackage{xcolor}
\newcommand{\yiq}[1]{{#1}}

\usepackage[capitalize,noabbrev]{cleveref}

\theoremstyle{plain}
\newtheorem{theorem}{Theorem}
\newtheorem{proposition}{Proposition}
\newtheorem{lemma}{Lemma}

\theoremstyle{definition}
\newtheorem{definition}{Definition}

\theoremstyle{remark}

\usepackage[textsize=tiny]{todonotes}

\icmltitlerunning{Semi-Offline Reinforcement Learning for Optimized Text Generation}

\begin{document}

\twocolumn[
\icmltitle{Semi-Offline Reinforcement Learning for Optimized Text Generation}




\begin{icmlauthorlist}
\icmlauthor{Changyu Chen}{yyy,inter}
\icmlauthor{Xiting Wang}{comp}
\icmlauthor{Yiqiao Jin}{yyy2}
\icmlauthor{Victor Ye Dong}{comp2}
\icmlauthor{Li Dong}{comp}
\icmlauthor{Jie Cao}{comp2}
\icmlauthor{Yi Liu}{comp2}
\icmlauthor{Rui Yan}{yyy}
\end{icmlauthorlist}

\icmlaffiliation{yyy}{Gaoling School of Artificial Intelligence, Renmin University of China, Beijing, China}
\icmlaffiliation{yyy2}{Georgia Institute of Technology, Atlanta, USA}
\icmlaffiliation{inter}{The work was done during the author’s internship at Microsoft Research Asia.}

\icmlaffiliation{comp}{Microsoft Research Asia, Beijing, China}
\icmlaffiliation{comp2}{Microsoft, Redmond, USA}


\icmlcorrespondingauthor{Xiting Wang}{xitwan@microsoft.com}
\icmlcorrespondingauthor{Rui Yan}{ruiyan@ruc.edu.cn}

\icmlkeywords{Machine Learning, ICML}

\vskip 0.3in
]


\printAffiliationsAndNotice{}

\begin{abstract}
In reinforcement learning (RL), there are two major settings for interacting with the environment: online and offline. Online methods explore the environment at significant time cost, and offline methods efficiently obtain reward signals by sacrificing exploration capability.   We propose semi-offline RL, a novel paradigm that smoothly transits from offline to online settings, balances exploration capability and training cost, and provides a theoretical foundation for comparing different RL settings. Based on the semi-offline formulation, we present the RL setting that is optimal in terms of optimization cost, asymptotic error, and overfitting error bound. Extensive experiments show that our semi-offline approach is efficient and yields comparable or often better performance compared with state-of-the-art methods. Our code is available at~\href{https://github.com/ChangyuChen347/semi-offline-RL}{https://github.com/ChangyuChen347/semi-offline-RL}.\looseness=-2
\end{abstract}

\begin{figure*}
  \includegraphics[width=1\textwidth]{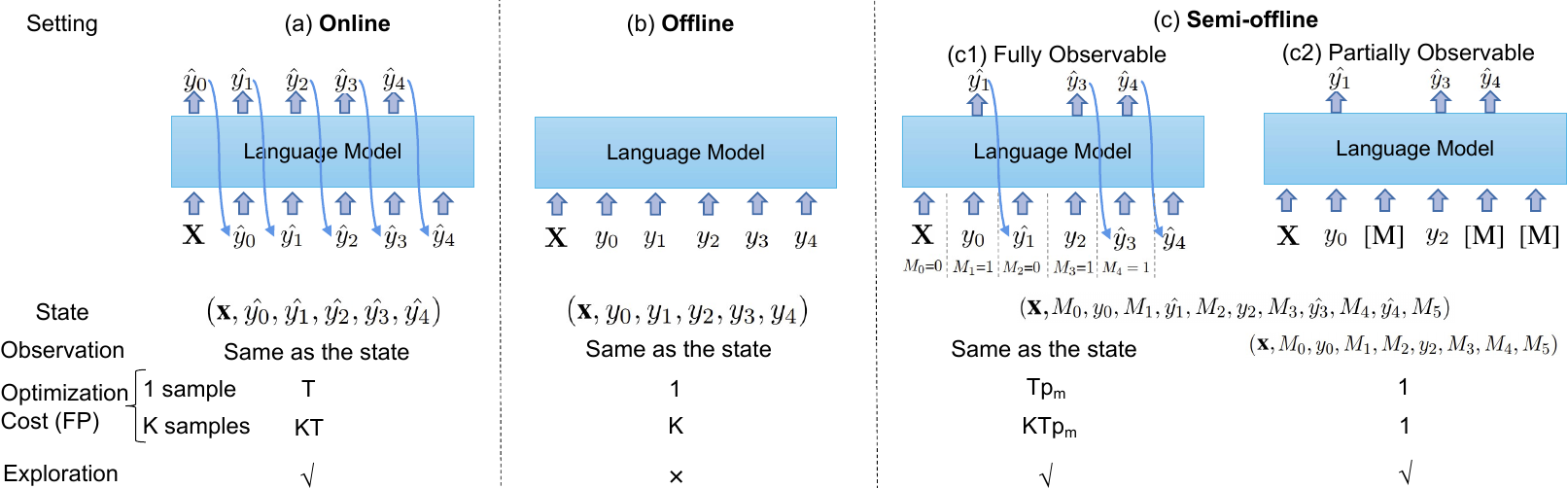}
  \vspace{-7mm}
  \caption{The comparison of different RL settings: (a) online methods explore the environment with a large optimization cost; (b) offline methods efficiently obtain reward signals by sacrificing the capability for exploration; (c) our proposed semi-offline setting enables exploration with minimum optimization cost. Here, $M_t=1$ (or $M_t=0$) means that the token at time $t$ is sampled based on the language model (or comes from static data), $0\leq p_m \leq 1$ is the probability for $M_t$ to be 1, FP denotes the number of forward propagations through the language model, and $T$ is the maximum number of word tokens in an output.
  }
  \vspace{-2mm}
  \label{fig:cmp_rl}
\end{figure*}

\section{Introduction}

Pretrained language models have achieved great success in improving text generation quality~\cite{devlin2019bert, liu2019roberta}. Recent research shows that a key for further improving pretrained language models is reinforcement learning (RL), which provides a principled solution for directly optimizing the final objective such as ROUGE~\cite{lin2003automatic}, factual correctness~\cite{goodrich2019assessing}, and human feedback~\cite{ouyangtraining}.
Recent large pretrained models that incorporate reinforcement learning, e.g., InstructGPT~\cite{ouyangtraining}, ChatGPT\footnote{~\url{https://openai.com/blog/chatgpt/}}, and GPT-4\footnote{~\url{https://openai.com/research/gpt-4}}, have demonstrated superior performance in aligning with user intent compared to GPT-3~\cite{zhao2023survey}.
In reinforcement learning, there are two major settings for interacting with the environment: online and offline. 

\textbf{Online RL} (Fig.~\ref{fig:cmp_rl}(a)).
The language model in this setting generates word token $\hat{y}_t$ by sampling from its output probability distribution, and obtains the reward signal about the samples to learn how well they fulfill the final objective~\cite{selfcritic,selfcriticranker,schulman2017proximal,Le2022CodeRLMC}.  
The online setting allows the language model to \textbf{fully explore} the environment: the model can interact with the environment to see the reward of different samples and hence obtains a comprehensive understanding about the final objective, which is crucial for finding the optimal generation.
While good generation can usually be found when the number of samples approaches infinity, it is empirically time-intensive to obtain even only a few samples from large pretrained language models. 
While there are many aspects of time cost in online RL~\cite{Wang2018ARL,Zhao2020LeveragingDF,Wang2022MultilevelRR,Feng2022ReinforcementRO,Yang2022ReinforcementSR},
this paper focuses on the forward propagations of language models. 
In particular, optimizing with $K$ samples requires $KT$ forward propagations (FPs) through the language models, where $T$ is the maximum number of tokens in the generated text.
This cost is quite large and impractical in some real-world scenarios~\cite{umpg} considering the complexity of large language models.

\textbf{Offline RL} (Fig.~\ref{fig:cmp_rl}(b)). This setting eliminates the need for generating text during the training process by utilizing a static dataset for learning.
Example static data $y_1,\cdots,y_T$ includes demonstrations or ground-truth labels~\cite{pang2020text,jaques2019way,serban2017deep,zhuscaling} for an input $\mathbf{x}$, as well as text pre-generated with beam search~\cite{liu2022brio}. 
By avoiding generating text in an autoregressive manner during training, offline methods \textbf{mitigate the expensive optimization costs} associated with online methods and reduces the cost from $KT$ forward propagations to $K$. 
However, offline methods \textbf{cannot explore} the environment to find the optimal generation: language models are only given the reward signals for specific static data, which prevents them from better understanding the final objective and converging to a better solution.

\yiq{
The above analysis shows that different RL settings have entirely different exploration capabilities and optimization costs. A fundamental research question is: \textbf{can we refine the RL setting so that effective exploration is achieved with minimum optimization cost}?
In this paper, we address this question by making three contributions.
}

First, we define the design space of different RL settings by proposing semi-offline RL, which bridges the gap between online and offline RL and provides a theoretical foundation to compare different RL settings.
As shown in Fig.~\ref{fig:cmp_rl}(c), semi-offline RL composes a sample by mixing tokens generated by the language model and tokens from the static dataset with a probability $p_m\in[0,1]$.
Different values of $p_m$ correspond to different MDPs allowing for a smooth transition from offline to online settings. 
In particular, for a fully observable scenario in which the model input (observation) is equal to the environment state (Fig.~\ref{fig:cmp_rl}(c1)), the semi-offline setting becomes offline when $p_m=0$, and becomes online when $p_m=1$.
When $p_m\in(0,1)$, we optimize the reward with an intermediate optimization cost $KTp_m$ while keeping the capability for exploring the environment.
Compared with the offline setting, semi-offline methods only utilize the static data as initial points for exploration, thereby allowing the model to identify better improvement directions.
Compared with the online setting, semi-offline methods may find the optimal improvement directions with a fewer number of samples by more quickly estimating token-wise rewards (Sec.~\ref{sec:opt}, Proposition~\ref{prop:tk}).

Second, based on the semi-offline MDP formulation, we present the RL setting that is optimal in terms of the following desirable properties: 
\begin{itemize}[nosep,leftmargin=1em,labelwidth=*,align=left]
\item[\textbf{DP1}.] Minimum optimization cost: the policy can be optimized by using only 1 FP per input.
\item [\textbf{DP2}.] Minimum asymptotic bias: the estimated error when the number of instances is unlimited is minimal among all possible RL settings that satisfy DP1.
\item [\textbf{DP3}.] Minimum overfitting error bound: the chosen RL setting has the lowest bound of error~\cite{franccois2019overfitting} when data is limited, among all settings that satisfy DP1 and DP2.
\end{itemize}

We prove that the optimal RL setting in terms of DP1$-$DP3 is easily implemented by mixing static data and the mask token, as shown in Fig.~\ref{fig:cmp_rl}(c2). 
This masked language model (MLM) setting fits naturally into existing pretrained language models, can explore $K$ samples with only 1 FP, and effectively find improvement directions by using static data points as a starting point.

Third, we evaluate our semi-offline RL approach in various text generation tasks and datasets, and show that it yields comparable or usually better performance compared to state-of-the-art methods while improving efficiency.

\yiq{
}





\section{Background}
\subsection{Preliminares about Reinforcement Learning}
In text generation, we often use a human-annotated corpus as ground truth for performing supervised learning. Reinforcement learning (RL) provides an additional way of learning in which the agent can optimize its behavior by interacting with the environment~\cite{Hyun2022GeneratingMS,Wang2022MultilevelRR,Chen2022PersonalizedCG}. An agent-environment interaction can be described by the following process: 1) the environment tells the agent the current \textbf{state}, 2) the agent outputs an \textbf{action} given the state through a function called \textbf{policy}, and 3) after the agent acts, the environment shifts to a new state and the environment gives a \textbf{reward} based on the agent's action and the updated state. The goal of the agent is to learn a policy that yields the maximum cumulative reward. We use a pretrained language model as the policy.\looseness=-1


\yiq{In RL, the environment is typically formulated as a \textbf{Markov Decision Process (MDP)}. An MDP is defined as a tuple $\textbf{M} = \{\mathcal{S}, \mathcal{A}, \mathcal{R}, \mathcal{T}\}$, where $\mathcal{S}$ and $\mathcal{A}$ denote the state space and action space, respectively. The reward function, $\mathcal{R}: \mathcal{S} \times \mathcal{A} \rightarrow \mathcal{R}$, maps a state-action pair to a scalar value, and the transition function, $\mathcal{T}: \mathcal{S} \times \mathcal{A} \times \mathcal{S} \rightarrow \mathcal{R}$, describes the probability of transitioning from one state-action pair to another. Given an MDP, various methods of RL can be applied in the search space of the environment to learn a policy that maximizes the cumulative reward.}

\subsection{Reinforcement Learning for Text Generation}


The MDP for text generation is usually defined as follows:

\textbf{State} $s_t \in \mathcal{S}$ consists of the input sequence $\textbf{x}$ and the part of output text that has already been derived: $s_t=(\textbf{x},y_0^s, \cdots, y_{t-1}^s)$. Here, $y_{t}^s \in \mathcal{V}$ is the output token at time $t$, and $\mathcal{V}$ denotes the vocabulary.

\textbf{Action} 
$a_t=y_{t}^s\in \mathcal{A}$ is one of the $|\mathcal{V}|$ tokens.

\textbf{Transition}: $\mathcal{T}(s_{t+1}|s_t, a_t)$ transits $s_t$ to $s_{t+1}$ at time step $t$: 
$s_{t+1}=s_t  \cup y_{t}^s$.
This deterministic transition appends the next token to the previous state.

\textbf{Reward}: 
$R(s_t)=f(\textbf{x},y_0^s, \cdots, y_{t-1}^s)$ quantifies how good the derived sentence $y_0^s, \cdots, y_{t-1}^s$ is according to the final objective like the BLEU score or user satisfaction.
In text generation, we often consider terminal reward, which means that the reward is computed after the whole text is generated, in other words $R(s_t)\neq 0$ only when $t=T$.

\label{sec:back}

Both online and offline RL methods can be depicted by using this MDP.

In the \textbf{online} setting, each action is obtained by sampling from the probability distribution (Fig.~\ref{fig:cmp_rl}(a)), i.e., $a_t=y_t^s=\hat{y}_t$, where $\hat{y}_t\sim p(\hat{y}_t|s_t;\theta)$, where $\theta$ is the parameter for the language model. Accordingly, the reward is computed by considering state $s_t=(\mathbf{x},\hat{y}_0,\cdots,\hat{y}_{t-1})$.
Online RL methods have a large search space, allowing them to search for the optimal solution across the entire space. However, this can make them difficult to optimize with high variance in reward signals. To address this, methods such as actor-critic~\cite{konda1999actor,bahdanau2016actor,Le2022CodeRLMC}, self-critic~\cite{zhang2019continuous,selfcritic,selfcriticranker}, and PPO~\cite{schulman2017proximal,ouyangtraining} have been developed.
However, these methods still explore the environment at great optimization cost due to the auto-regressive generation of output text. 

In the \textbf{offline} setting, each action is derived by using the token in the static data (Fig.~\ref{fig:cmp_rl}(b)), i.e., $a_t=y_t^s=y_t$, where $y_t$ is the $t$-th token in the static data.
Accordingly, the reward is computed by considering  state $s_t=(\textbf{x},y_0,\cdots,y_{t-1})$.
RAML ~\cite{norouzi2016reward} can be viewed as one of the pioneering offline RL methods. It obtains the static dataset for offline learning by edit-distance sampling and weighs the samples according to the reward, resulting in a formulation of reward augmented maximum
likelihood. SPG ~\cite{spg} and ERPO ~\cite{erpo} improve upon RAML~\cite{norouzi2016reward} by exploring a larger region but introducing more decoding cost and they fall into the category of online methods.
Offline methods are also widely used in dialogue systems to reduce the number of interactions with people in real-time~\citet{serban2017deep,jaques2019way}.
Recently, GOLD~\cite{pang2020text} posits that acquisition of useful data through exploration can be challenging. Therefore, it directly employs the ground-truth. 
BRIO~\cite{liu2022brio}, on the other hand, harnesses a contrastive loss and generates multiple candidates for each instance as the static dataset.
Unlike previous methods, ILQL~\cite{snell2022offline} involves training a value network on a static dataset. During deployment, this value network is used to perturb the output distribution of another language model trained with supervised learning.
While offline methods efficiently obtain reward signals by leveraging the static dataset, they sacrifice the exploration capability.

\section{Semi-Offline MDP} 
\label{sec:semioffline}

\subsection{Formulation of Semi-Offline MDP}
\label{sec:form}
\yiq{
In order to lay a theoretical foundation for comparing different RL settings, we define the design space of different RL settings by proposing semi-offline RL, which can smoothly transit from offline methods to online methods by using different values of hyperparamter $p_m$.
More specifically, semi-offline RL composes a sample by mixing tokens generated by the language model and tokens from the static dataset with a probability $p_m\in[0,1]$, as shown in Fig.~\ref{fig:cmp_rl}(c). 
The formal definition is as follows.
}
 \begin{definition}[MDP of semi-offline RL]
In semi-offline RL:

\textbf{State} $s_t=(\textbf{x},M_0, y_0^s, \cdots, M_{t-1}, y_{t-1}^s, M_t)$ consists of the input sequence $\textbf{x}$, the part of output text that has already been derived $y_0^s,\cdots,y_{t-1}^s$, as well as the binary values $M_0,\cdots,M_t$, each $M_t\in \{0, 1\}$ denotes whether the next token $y^s_t$ will be determined according to the agent's generation ($M_t=1$) or the static dataset ($M_t=0$). 

\textbf{Action} 
$a_t$ is the output token of agent at time $t$. If $M_t=1$, the agent will output one of the $|\mathcal{V}|$ tokens by sampling from the probability distribution of the language model: $a_t=\hat{y}_{t}$. If $M_t=0$, The agent will give a $NULL$ token.
  
\textbf{Transition}: $\mathcal{T}(s_{t+1}|s_t, a_t)$ transits $s_t$ to $s_{t+1}$ at time $t$ with\looseness=-1
\begin{align}
s_{t+1}&=s_t  \cup y_{t}^s \cup M_{t+1}, \\
y_{t}^s&=
\begin{cases}
\hat{y}_{t}, & \text{if } M_t=1\\
{y}_{t}, & \text{if } M_t=0
\end{cases} \\
M&_{t+1} \sim Bernoulli(p_m)
\end{align}

where $\hat{y}_{t}$ is a token generated by the language model, $y_{t}$ is a token from the dataset (e.g., the $t$-th token in the ground-truth), and $M_{t+1}$ is sampled from a Bernoulli distribution parameterized with $p_m$, which means that $M_{t+1}$ takes the value of 1 with a probability $p_m$ and takes the value 0 with a probability $1-p_m$.

\textbf{Reward}: 
$R(s_t)=f(\textbf{x},y_0^s, \cdots, y_{t-1}^s)$ quantifies how good the generated sentence $y_0^s, \cdots, y_{t-1}^s$ is according to the ultimate goal like the BLEU score or user satisfaction.

\label{def:1}
\end{definition}

Semi-offline RL is most comparable to online and offline settings in the \textbf{fully observable} scenario shown in Fig.~\ref{fig:cmp_rl}(c1).
Fully observable means that the model input (observation) is equal to the state of the environment, meaning that the language model is aware of all information in the state when making a decision, i.e., $\hat{y}_t\sim p(\hat{y}_t|s_t;\theta)$.

In this fully observable scenario, $p_m=0$ is equivalent to the offline setting where the model optimizes its policy from a completely static dataset. Conversely, when $p_m=1$, the configuration is in the online mode, allowing for dynamic exploration of the maximum search space. When $p_m\in(0,1)$ is an intermediate value, the semi-offline MDP balances between dynamic exploration of the search space and knowledge obtained from the static dataset, while also finding an equilibrium between exploration and time cost.
More specifically, the time cost for semi-offline methods can be computed with the following proposition.



\begin{proposition}[Time cost when fully observable]
Considering the fully observable scenario in which all information in the states is observed by the language model to get sampled tokens. Let us denote the minimum number of FPs required to sample $s_t$ as $C_t$ and its expectation as $\mathbb{E}(C_t)$. We have
\begin{equation}
    C_t=\sum_{t'=0}^{t-1}M_{t'},  \textnormal{       }
    \mathbb{E}(C_t)=tp_m
\end{equation}
\label{prop:1}
\end{proposition}
The proof is given in Appendix~\ref{appendix:a1}. Proposition~\ref{prop:1} shows that when $p_m\in(0,1)$, we could optimize the reward with an intermediate optimization cost controlled by $p_m$ while keeping the capability for exploring the environment.

In addition to the time cost, the intermediate methods whose $p_m\in(0,1)$ provides an additional view for exploration.
Compared with the offline setting, semi-offline methods only utilize the static data as initial points for exploration instead of seeing only the reward of static data points, thereby allowing the model to get a more comprehensive understanding about the final objective and identify better improvement directions.
Compared with the online setting, semi-offline methods may find the optimal improvement directions with a fewer number of samples.
This sampling efficiency is achieved by only exploring a vicinity of the static data point.
Thus, the space to be explored for semi-offline methods ($|\mathcal{V}|^{Tp_m}$) is exponentially smaller than that of the online methods ($|\mathcal{V}|^{T}$), making it easier for the language model to understand the reward gain brought by different choices.
Even though the exploration space is limited, it is possible that the knowledge explored in the vicinity of specific output text can be generalized to other output text considering the generalization ability of neural networks. This is verified by our experiments, which show that semi-offline usually performs equally good or better with much less time cost compared with existing online or offline methods (Sec.~\ref{sec:exp}).\looseness=-1 
\subsection{RL Setting with Minimum Optimization Cost}
Next, we move towards achieving the minimum optimization cost while maintaining the effective exploration capability of the agent.
In particular, we are interested in finding a semi-offline RL setting that can be optimized with only 1 FP per instance, so that even large pretrained language models could be optimized efficiently.
Meanwhile, we hope that the agent can still freely decide the degree for exploration by choosing different values of $p_m$.\looseness=-1

We can see from Propsition~\ref{prop:1} that the time cost in the fully observable scenario cannot always be 1 FP for different values of $p_m$. In order to further accelerate the optimization method, we must remove the condition of full observation, i.e. not requiring $a_t$ to be a decision made after observing all information in $s_t$.
This scenario can be formulated by using Partially Observable Markov Decision Process (POMDP).
\begin{definition}[Semi-offline MDP when partially observable]
The MDP of the environment is the same as that in Def.~\ref{def:1}.
However, the agent in POMDP takes action $a_t$
based on observation $o_t$, which does not contain all information in $s_t$:
\begin{equation}
a_t =\hat{y}_{t}\sim p(\hat{y}_{t}|o_t;\theta), \textnormal{\ 
\ when    } M_t=1 \\    
\end{equation}
$o_t$ is a sub-sequence of $s_t=(\textbf{x},M_0,y_0^s, M_1 \cdots, y_{t-1}^s, M_{t})$.
\label{def:2}
\end{definition} 

Losing information in $s_t$ may significantly decrease the probability of achieving optimal results. 
To derive an optimal RL setting under the minimum time cost constraint, we consider two research questions: 
\begin{itemize}[nosep,leftmargin=1em,labelwidth=*,align=left]
\item[\textbf{RQ1}.] Which information has to be removed from the observation in order to meet the minimum time cost constraint?
\item [\textbf{RQ2}.] What information needs to be retained in the observation to maximize the performance?  
\end{itemize}

RQ1 can be answered with the following proposition.
\begin{proposition}[Information loss with minimum time cost]
If $s_{T}$ can always be sampled within 1FP for $\forall p_m\in[0,1]$, then $o_t$ must \textbf{not} contain any exact information about sampled tokens $\hat{y}_{t'}$, for $\forall t' \in [0, t-1]$ and $\forall t \in [1, T)$.
\label{prop:2}
\end{proposition}
The proof is given in Appendix~\ref{appendix:a2}.
This proposition can be easily understood: 
when aiming for parallel generation of different tokens, the generation of one token $\hat{y}_t$ should not rely on the information of another token $\hat{y}_{t'}$ generated simultaneously.

Proposition~\ref{prop:2} allows us to define the maximum set of observations we can get at time step $t$ when minimum time cost can be achieved, which is important for answering RQ2.
\begin{definition}[Maximum observation with minimum time cost]
If $s_{T}$ can always be sampled within 1FP for $\forall p_m\in[0,1]$, the observation with the maximum information in $s_t$ is
$o_t^{max}=(\textbf{x}, M_0, y^o_0,\cdots, M_{t-1}, y^o_{t-1}, M_t )$,
where
\begin{equation}
y^o_t=\left\{
\begin{array}{rcl}
NULL & & M_t=1\\
y_t & & M_t=0\\
\end{array} \right.
\end{equation}    
\label{def:maximum_observation}
\end{definition}

With Def.~\ref{def:maximum_observation}, we can answer RQ2 by characterizing the asymptotic bias of RL methods, which refers to the error of an agent with unlimited data. This characterization is achieved by using the following lemma.

\begin{lemma}[Criteria for 0 asymptotic bias] According to Theorem 1 and Definition 2.4 in~\cite{franccois2019overfitting}, the additional error introduced by changing $o_t^{max}$ to $o_t$ when the data is unlimited is 0, if for $\forall t$ and $\forall s$
\begin{equation}
p(s|o_t) = p(s|o_t^{max})
\end{equation}  
\label{lemma:asymptotic bias}
\end{lemma}

Another measure for performance is the overfitting error, which depicts the additional error introduced due to limited data. The following lemma shows the criterion for achieving minimum overfitting error bound.
\begin{lemma}[Criteria for minimum overfitting error bound]    
Accordingly to Theorem 3 in~\cite{franccois2019overfitting}, the overfitting error bound is minimized when the number of possible observations $|O_t|$ is minimized, where each $o_t\in O_t$ is an element in set $O_t$. 
\label{lemma:overfit}
\end{lemma}


Lemmas~\ref{lemma:asymptotic bias} and~\ref{lemma:overfit} provide a theoretical foundation to decide whether a RL setting can achieve optimal performance.
Lemma~\ref{lemma:asymptotic bias} shows that all information useful for predicting $s$ should be kept in the observation, and Lemma~\ref{lemma:overfit} claims that the observation should contain as little redundant information as possible to avoid overfitting. 
This allows us to rule out methods such as Scheduled Sampling~\cite{bengio2015scheduled,mihaylova2019scheduled}, which does not contain the information of $M_t$ and thus cannot satisfy Lemma~\ref{lemma:asymptotic bias}. They also help exclude observations that include additional information such as $y_t$ when $M_t=1$, which fails to satisfy Lemma~\ref{lemma:overfit}.

According to these lemmas, we define optimal RL setting under the minimum time cost constraint as follows:



\begin{definition}[Optimal RL setting]
In the optimal RL setting, its observation $o^*_t$ should satisfy
\begin{itemize}[nosep,leftmargin=1em,labelwidth=*,align=left]
\item[DP1.] Minimum time cost: $s_T$ can always be sampled within 1 FP.
\item[DP2.] Minimum asymptotic bias: $o^*_t$ satisfies the criteria for 0 asymptotic bias given by Lemma~\ref{lemma:asymptotic bias}.
\item[DP3.] Minimum overfitting error bound: $o^*_t$ satisfies the minimum overfitting error bound criteria in Lemma~\ref{lemma:overfit}.
\end{itemize}
\label{def:3}
\end{definition}


We then prove that the optimal RL setting in Def.~\ref{def:3} could be easily implemented by mixing static data and the mask token, as shown in Fig.~\ref{fig:cmp_rl}(c2), where [M] denotes a mask token. 
This masked observation setting fits naturally into existing pretrained language models and can explore $K$ samples with only 1 FP.
Formally, we define the RL setting with masked observations as follows. 
\begin{definition}[RL setting with masked observations]
The masked observation is defined as 
    $o_t^M$=$x,y_0^M,y_1^M,\cdots,y_{t-1}^M$
    where
    \begin{equation}
y^M_t=\left\{
\begin{array}{rcl}
\text{[M]} & & M_t=1\\
y_t & & M_t=0\\
\end{array} \right.
\end{equation}
\label{def:4}
\end{definition}

We then formally prove the optimality of masked observation with the following theorem. 
\begin{theorem}[Optimality of masked observation]
    $o^M_t$ in Def.~\ref{def:4} is $o^*_t$ in Def.~\ref{def:3}.
\label{theorem:mask}
\end{theorem}
The proof of Theorem~\ref{prop:1} is given in Appendix~\ref{appendix:a3}.

\subsection{Optimization}
\label{sec:opt}
\subsubsection{RL Loss for Solving MDP}
POMDP defined in Def.~\ref{def:4} can be solved in the same way as traditional MDPs. Here we use policy gradient because pretrained language models provide a natural initialization of the policy.
Specifically, we adopt the REINFORCE with baseline~\cite{williams1992policygradient} to reduce the variance among different trajectories.
Accordingly, the policy is optimized with the following RL loss:
 \begin{equation}
\begin{array}{rcl}
& \mathcal{L}_{RL}=\frac{1}{K}\sum_{k=1}^K - (\mathcal{R}(Y^k)-b) \sum_t \log\ p(a_t^k|o^M_t)  \\
 &   b=\frac{\sum_k\mathcal{R}(Y^k)}{K}
\end{array} 
\label{eq:rl_f}
\end{equation}
where $K$ is the number of samples, $k$ denotes the sample index, $Y^k=(a^k_0, \cdots, a^k_{T-1})$ is the $k$-th sampled sentence, and $b$ is the baseline computed by averaging the rewards of sampled sentences to reduce the variance. 

\textbf{Analysis of optimization cost}. We can easily see that the number of FPs needed for computing $\mathcal{L}_{RL}$ is always 1, regardless of the number of samples. This is because different samples are obtained by using the same observation $o^M_t$, and thus can be obtained together with 1 FP. 

\textbf{Efficient estimation of token-wise rewards}.
Decomposing $\mathcal{L}_{RL}$ into token-wise rewards using the following proposition allows us to see that the RL setting with masked observations enables more efficient learning of token-wise rewards.\looseness=-1

\begin{proposition}[Token-level reward assignment]
\label{prop:tk}
$\mathcal{L}_{RL}$ in Eq.~\ref{eq:rl_f} can be decomposed into token-wise loss $\mathcal{L}_{t}^i$ of the $i$-th token in the vocabulary at time step $t$: 
\begin{equation}
\label{prop:4}
\begin{array}{lll}
\mathcal{L}_{RL} = \sum_{t} \sum_{i=1}^{|\mathcal{V}|}\mathcal{L}_{t}^i\\
\mathcal{L}_{t}^i= -\frac{C_t^i}{K}\ \log \ 
p(\mathcal{V}_i|o_t^M) (\frac{\sum_{k=1, \mathcal{V}_i=a_t^k}^K  \mathcal{R}(Y^k)}{C_t^i}-b) \\
b=\frac{\sum_{k=1}^K\mathcal{R}(Y^k)}{K} \\
\end{array}
\end{equation}

\end{proposition}


where $\mathcal{V}_i$ is the $i$-th token in the vocabulary, and $C^i_t$ denotes the number of samples that select $\mathcal{V}_i$ as the action at time step $t$. $a_t^k$ is the $t$-th token of output $Y^k$.

The proof is in Appendix~\ref{appendix:a4}.
Proposition~\ref{prop:tk} shows that we evaluate how well the token $\mathcal{V}_i$ performs at time $t$ by computing $\frac{\sum_{k=1, \mathcal{V}_i=a_t^k}^K  \mathcal{R}(Y^k)}{C_t^i}-\frac{\sum_{k=1}^K\mathcal{R}(Y^k)}{K}$, which estimates $\mathbb{E}_{Y \sim p(Y|o=o^M_t,a=\mathcal{V}_i)}\mathcal{R}(Y)-\mathbb{E}_{Y \sim p(Y|o=o_t^M)}{\mathcal{R}(Y)}$, the expected advantage of generating $\mathcal{V}_i$ under observation $o_t^M$. 

The accurate estimation of this expected advantage requires a large number of samples under the same observation $o$. This can be easily achieved in our semi-offline RL setting with masked observations as $o_t^M$ remains constant for different sampled tokens. In comparison, observation $o$ is usually different for different samples in the online and offline RL setting. Thus, they may require more samples to accurately understand the contribution of a single token $\mathcal{V}_i$.
\subsubsection{Overall Optimization Loss}
We follow existing paradigm for RL training. Specifically, pretrained language models are first fine-tuned with the ground-truth labels to ensure a good starting point for RL training. This is achieved by optimizing the MLE loss 
\begin{equation}
    \mathcal{L}_{MLE} = \sum_{t=1}^{|y^{gt}|} \log\ p(y_{t}^{gt}|x,y_1^{gt},,\cdots,y_{t-1}^{gt})
\end{equation}
where $y_{gt}$ is the ground-truth in the dataset.
During this phase, we replace some tokens in the input $y_1^{gt},\cdots,y_{t-1}^{gt}$ with [M] so that the model can be better adapt to masks in the inputs.

We then perform RL training by simultaneously considering both the MLE loss and the RL loss. The MLE loss is considered here to prevent the policy from drifting away from the original dataset, which may lead to a reduction in generation quality.
This is achieved by minimizing
\begin{align}
    \mathcal{L} = \mathcal{L}_{MLE}+\lambda \mathcal{L}_{RL}
\end{align}
where $\lambda>0$ is a hyperparameter.

\begin{table*}[t]
\caption{Overall performance. FP denotes the number of forward propagations required for optimization, $N$ is number the instances consumed by the model, $K$ is the number of samples used, and $L$ is the sentence length.}
\label{main}
\vskip -0.1in
\begin{center}
\begin{small}
\begin{sc}
\resizebox{1\linewidth}{!}{
\begin{tabular}{lllccccccccccccc}
\toprule
\multirow{2}{1cc}{{Groups}} & \multirow{2}{1cc}{{Models}} & \multirow{2}{1cc}{{FP}} & \multicolumn{3}{c}{CNN/DM} &
\multicolumn{3}{c}{SAMSum} &
\multicolumn{3}{c}{SQuAD} &
\multicolumn{3}{c}{XSum} &
\\
\cmidrule(lr){4-6}  \cmidrule(lr){7-9} \cmidrule(lr){10-12}  \cmidrule(lr){13-15} 
& &  & R-1 & R-2 & R-l &  R-1 & R-2 & R-L & B-4 & R-L & MTR & R-1 & R-2 & R-L\\ 
\midrule
\multirow{2}{1cm}{Base} & FT & N  & 44.16 & 21.28 & 40.90 & 53.32 & 28.53 & 49.03 &
27.21 & 54.13 & 27.70&
47.46 & 24.69 & 39.53\\
& M-FT & 
N & 45.10 & 21.76 & 41.86 & 53.09 & 28.17 & 49.02 & 
27.43 & 54.30 & \underline{27.82} &
47.65 & 24.85 & 39.56\\
\midrule
\multirow{2}{1cc}{Offline} 
& GOLD & 
N & 45.51 & 22.10 & 42.30 &
53.18 & 28.90 & 49.11 &
27.20 & 54.43 & 27.59 &
47.75 & 24.92 & 39.70\\
& BRIO & 
NK & 47.83 &23.75 & 44.65 & 53.98 & 29.11 & 49.56&
27.17 & 54.53 & 27.64&
\underline{48.91} & \textbf{25.71} & \textbf{40.60} \\
\midrule
\multirow{3}{1cc}{Online} & SC & NT & 45.45 & 21.85 & 42.16 & 53.47 & 28.54 & 48.99&
27.14 & 54.36 & 27.58 &
47.90 & 24.95 & 39.73 \\
& AC & NT & 45.71 & 22.07 & 42.42  & 53.41 & 28.29 & 48.90 &
27.35 & 54.48 & 27.62 &
47.88 & 24.92 & 39.71 \\
& AVG &NTK & \underline{48.28} & \underline{24.16} & \underline{45.00} & \underline{54.10} & \textbf{29.21} & \underline{49.58} &
\underline{27.50} & \underline{54.79} & 27.77 &
48.48 & 25.21 & 40.23  \\
\midrule
Semi & Ours & N & \textbf{48.54} & \textbf{24.40} & \textbf{45.35}  & \textbf{54.27} & \underline{29.19} & \textbf{50.57}&
\textbf{27.79} & \textbf{54.95} & \textbf{28.32} &
\textbf{49.02} & \underline{25.37} & \underline{40.52}  \\
\bottomrule
\end{tabular}
}
\end{sc}
\end{small}
\end{center}
\vskip -0.1in
\end{table*}

\section{Experiment}
\label{sec:exp}
\subsection{Experimental Setup}

\subsubsection{Datasets}
We conduct experiments on 1) a summarization dataset \textbf{CNN/DM}~\cite{cnn}, where the goal is to generate summaries for news articles; 2) a dialogue summarization dataset \textbf{SAMSum}~\cite{gliwa-etal-2019-samsum}, in which the focus is summarizing dialogues instead of news articles; 3) a natural question generation dataset \textbf{SQuAD}~\cite{SQuAD}, where the task is to generate questions that can be answered by a specific segment of an article; 4) an extreme summarization dataset \textbf{XSum}~\cite{narayan2018don}, which focus on generating highly abstractive summaries for news articles from the BBC. More statistical information about these datasets can be found in Appendix~\ref{appendix:c}. We have also experimented with other tasks such as advertisement generation, and the results can be found in Appendix~\ref{appendix:other_task}.





\subsubsection{Compared Methods}
\label{sec:cmp}
\textbf{Base models}.
We fine-tune (\textbf{FT}) pre-trained language models such as BART~\cite{lewis2020bart} and T5~\cite{t5} for each task. 
Specifically, for CNN/DM and SAMSum we use BART-large (406M). For SQuAD and XSum we use T5-large (737M) and Pegasus-large (570M)~\cite{zhang2020pegasus} respectively.

Additionally, we fine-tune the tasks with masks (\textbf{M-FT}) to study the influence of involving masks during training. This method is similar to FT but with the added step of randomly masking a portion of the tokens in the targets, giving the model the ability to predict the next token when given the special token [M] on these downstream tasks.

\textbf{Online methods}.
The online methods we compare include the online generation of single-sample methods Self-Critic (\textbf{\textsc{SC}})~\cite{selfcritic,selfcriticranker} and Actor-Critic (\textbf{\textsc{AC}})~\cite{Le2022CodeRLMC}.  
Both methods are optimized using REINFORCE with baseline~\cite{sutton1999policy}, where the baseline for \textsc{SC} is the greedy decoding result of the agent, and \textsc{AC} uses a quality scoring critic model to compute the reward.  
Average baseline (\textbf{\textsc{AVG}}) is a multiple-sample approach, in which its RL loss using the average rewards of the multiple samples as the baseline. The RL loss of AVG is also a variant of the contrastive loss used by BRIO~\cite{liu2022brio}\footnote{The proof of deriving BRIO to AVG can be found in Appendix~\ref{appendix:d}.}.

\textbf{Offline methods}.
The offline methods we compare are \textbf{\textsc{GOLD}}~\cite{pang2020text} and \textbf{\textsc{BRIO}}~\cite{liu2022brio}, where \textsc{GOLD} uses the original ground truth as the static dataset in offline training, and \textsc{BRIO} uses the generated results of the \textsc{Base} model as the static dataset.

To ensure a fair comparison, we employ M-FT as the initialization method and keep the base model the same for ours and all RL methods. For BRIO and ours, we use the same static dataset. For more implementation details of how we collect trajectory for online, offline, and our semi-offline method, one can refer to Appendix~\ref{appendix:b}.

\subsubsection{Metrics}
Following~\cite{liu2022brio,bengio2015scheduled}, We use ROUGE scores including \textbf{\textsc{R-1}}, \textbf{\textsc{R-2}}, and \textbf{\textsc{R-L}}~\cite{lin2003automatic} to evaluate the results on the summarization tasks: CNN/DM, SAMSum, and XSum.
For SQuAD, we follow~\cite{ushio-etal-2022-generative} and adopt the BLEU score \textbf{\textsc{B-4}}~\cite{papineni2002bleu}, ROUGE scores~\cite{lin2003automatic}, and METEOR score \textbf{\textsc{MTR}}~\cite{Banerjee2005METEORAA}. 
We directly use the corresponding metrics as the reward to be optimized. For more implementation details of the reward and other hyperparameters, one can refer to Appendix~\ref{appendix:b}.
In addition to these metrics, we have also tested how our method performs when optimizing other metrics such as factuality (Appendix~\ref{appendix:other_task}) and whether humans consider the generated text to be better (Appendix~\ref{appendix:human}).

\subsection{Overall Performance}

Tab.~\ref{main} shows the overall performance of different models as well as their optimization cost (FP).
We have three main observations by analyzing the table.

First, our performance is significantly higher compared with other methods with the same FPs (i.e., FP=$N$). This is because that our method is the only one that has the exploration capability when FP is $N$. Other methods that have the same optimization cost are either base models or offline methods that only consider one sample.

Second, methods that only utilize one sample usually performs worse than multi-sample methods, even when they are time-expensive online methods such as \textsc{SC} and \textsc{AC}). 
This demonstrates the necessity to obtain more reward signals by exploring the environment. Our method is the only one that can increase the number of samples without increasing the number of FPs required for optimization.

Third. our method performs as well as or better than state-of-the-art methods that are much more costly than ours (e.g., \textsc{AVG}).
We achieve the best result on three datasets, demonstrating the superiority of our semi-offline setting in addition to the significantly reduced optimization cost.
Although our result in \textsc{XSum} is only similar to \textsc{BRIO}, we show that our method still brings additional benefits. More specifically, Tab.~\ref{XSum:combing} shows that combining our method with \textsc{BRIO} results in improved performance.

We also provide human evaluation and results on more metrics in Appendix~\ref{appendix:human}.

\begin{table}[t]
\caption{The results of combining the loss of \textsc{BRIO} and \textsc{Ours}.}
\label{XSum:combing}
\vskip -0.05in
\begin{center}
\begin{small}
\begin{sc}
\resizebox{0.65\linewidth}{!}{
\begin{tabular}{lccc}
\toprule
\multirow{2}{1cm}{{Models}} & \multicolumn{3}{c}{XSum} \\
\cmidrule(lr){2-4}  
&  R-1 & R-2 & R-l \\
\midrule
BRIO & 48.91 & \underline{25.71} & \underline{40.60}  \\
Ours & \underline{49.02} & 25.37 & 40.52 \\
BRIO+Ours & \textbf{49.23} & \textbf{25.98} & \textbf{41.01}\\
\bottomrule
\end{tabular}
}
\end{sc}
\end{small}
\end{center}
\vspace{-4mm}
\end{table}

\begin{table*}[t]
\caption{Ablation study on variants that do not satisfy Lemma~\ref{lemma:asymmetric bias} or Lemma~\ref{lemma:overfit}.}
\label{ablation}
\vskip -0.1in
\begin{center}
\begin{small}
\begin{sc}
\resizebox{1\linewidth}{!}{
\begin{tabular}{lcccccccccccc} 
\toprule
\multirow{2}{*}{Models} & \multicolumn{3}{c}{CNN/DM} & \multicolumn{3}{c}{SAMSum} & \multicolumn{3}{c}{SQuAD} & \multicolumn{3}{c}{XSum} \\ 
\cmidrule(lr){2-4}\cmidrule(lr){5-7}\cmidrule(lr){8-10}\cmidrule(l){11-13} 
 & R-1 & R-2 & R-l & R-1 & R-2 & R-l & B-4 & R-L & MTR & R-1 & R-2 & R-l \\ 
\midrule
Ours & 48.54 & 24.40 & 45.35 
& 54.27 & 29.19 & 50.57  
& 27.79 & 54.95 & 28.32 
& 49.02 & 25.37 & 40.52 \\ 
\midrule
-Mask & 47.95 & 24.00 & 44.85 
& 54.19 & 28.65 & 50.23  
& 27.78 & 54.92 & 28.28 
& 48.80 & 25.34 & 40.40 \\
-Mask, $p_m$=1 & 47.43 & 23.67 & 44.48 
& 53.64 & 28.50 & 50.16 
& 27.57 & 54.94 & 28.07 
& 48.88 & 25.37 & 40.49 \\
+Noisy Mask & 48.14 & 24.03 & 45.03 
& 54.01 & 29.07 & 50.22 
& 27.50 & 54.91 & 28.00 
& 48.55 & 25.27 & 40.34 \\
\midrule
+ALL & 47.90 & 23.98 & 44.82
& 53.95 & 28.89 & 50.04
& 27.65 & 54.74 & 27.98 
& 48.71 & 25.22 & 40.41 \\
+PRE & 48.34 & 23.86 & 45.30 
& 54.04 & 29.00 & 50.17 
& 27.68 & 54.71 & 28.10 
& 48.44 & 25.13 & 40.17 \\
\bottomrule
\end{tabular}
}
\end{sc}
\end{small}
\end{center}
\vskip -0.1in
\end{table*}

\subsection{Optimization Efficiency}
To fully evaluate the performance of our method, it is important to consider not only the number of FPs, but also the real time cost during optimization. To this end, we fix the number of instances and compared the optimization speed as well as model performance in Tab.~\ref{SQuAD:time}. The experiments are run on a machine with an Nvidia A40 GPU (memory: 48 GB) using a learning rate of 1e-6 and a batch size of 8 for all compared methods. The results show that our method not only has the lowest training time consumption, but also the best optimization speed on both the SQuAD and SAMSum datasets. 
\begin{table}[h]
\vskip -0.1cm
\caption{Optimization efficiency on SQuAD and SAMSum. Time is measured in minutes. }
\label{SQuAD:time}
\vskip -0.7cm
\begin{center}
\begin{small}
\begin{sc}
\resizebox{1\linewidth}{!}{
\begin{tabular}{lcccrcr}
\toprule
\multirow{2}{1cc}{{Models}} & \multirow{2}{1cm}{{$\#$data}} & \multicolumn{3}{c}{SQuAD} & \multicolumn{2}{c}{SAMSum}\\
\cmidrule(lr){3-5}  \cmidrule(lr){6-7}
& & B-4 & R-L & time  & R-L & time\\ 
\midrule
\multirow{2}{1cm}{Ours} & 8K & 27.66 & 54.80 &  14.7 & 49.75 & 8.9\\
& 16K & 27.64 & 54.75 & 29.3 & 49.96 & 18.0\\
\midrule
\multirow{2}{1cm}{BRIO} 
& 8K & 27.42 & 54.36 & 18.8 & 49.39 & 9.3\\
& 16K & 27.50 & 54.46 & 37.5 & 49.35 & 19.0\\
\midrule
\multirow{2}{1cm}{Avg} 
& 8K & 27.62 & 54.70 & 121.0 & 49.09 & 135.8\\
& 16K & 27.58 & 54.72 & 243.0 & 49.49 & 271.0 \\

\bottomrule
\end{tabular}
}
\end{sc}
\end{small}
\end{center}
\vskip -0.25in
\end{table}
It is also worth noting that BRIO and AVG use multiple target texts for the same source text for one instance, which leads to increased \textbf{memory usage} on GPUs. In contrast, our method is more memory efficient as it only uses one target for each instance.

\subsection{Ablation Study}


This ablation study investigates 1) the impact of using an MDP that does not satisfy Lemma~\ref{lemma:asymmetric bias} or Lemma~\ref{lemma:overfit}, and 2) the effect of using different offline datasets for training.

\subsubsection{Design of MDP: Lemma1}
As mentioned in Sec.\ref{sec:semioffline}, failing to satisfy Lemma~\ref{lemma:asymmetric bias} will result in suboptimal results. 
We evaluate the correctness of this statement by devising the following variants:

1. \textbf{\textsc{-Mask}}: remove the mask information, and the $p_m$ of the environment is consistent with the main experiment (0.4). In this variant, we obtain the trajectory by Scheduled Sampling with a fixed choosing probability of 0.4~\cite{bengio2015scheduled,mihaylova2019scheduled}; 

2. \textbf{\textsc{-Mask, $p_m$=1}}: also remove the mask information and $p_m$ of the environment is set to 1;

3. \textbf{\textsc{+Noisy Mask}}: with mask information included, the model receives part of the wrong mask information, i.e. the environment tells the model that $M_t=0$ when in fact the 
environment's $M_t=1$;


As shown in Tab.~\ref{ablation}, All methods that violate Lemma~\ref{lemma:asymmetric bias} have a certain degree of performance drop. \textsc{+Noisy Mask} partially removes the mask information while \textsc{-Mask} and \textsc{-Mask, $p_m$=1} completely remove the mask information. Without mask information, the model is unable to account for how the environment mixes the dataset and model predictions, leading to inaccurate estimation of the reward for current actions, resulting in a large variance of the reward signal and poor optimization results. 



\subsubsection{Design of MDP: Lemma2}
For Lemma~\ref{lemma:overfit}, we design two new baselines that incorporate additional information by adding the complete sequence of the offline static data, denoted as $Y^d$, to the current observation, represented as $o_t=o_t \cap Y^d$.
. We empirically validate if this will result in overfitting and a drop in results on the test set. 

1. \textbf{+ALL}: we replace the input source $X$
 in the encoder as a concatenated sequence $X$ + [END] + $Y^d$. 

2.\textbf{+PRE}: we modify the decoder input by adding the embedding of the corresponding token in the target sequence $Y^d_i$
 to the embedding of the token [M] at the $i$-th position, i.e., Embedding([M]) is replaced with Embedding([M])+Embedding($Y^d_i$).

If we ignore the difference in input method of $Y^d$, $|O_t|$ of \textsc{+ALL} is larger than that of \textsc{+PRE}: for the $t$-th position, \textsc{+ALL} can see the full sequence $Y^d=(Y^d_1, \cdots, Y^d_{|Y^d|})$, while \textsc{+PRE} can only see the prefix $(Y^d_1, \cdots, Y^d_{t})$. The results from Tab.~\ref{ablation} show that \textsc{+ALL} and \textsc{+PRE} are both worse than ours. 



\begin{table*}[t]
\caption{Performance of using different static datasets. 
}
\label{cond2}
\vskip 0in
\begin{center}
\begin{small}
\begin{sc}
\resizebox{1\linewidth}{!}{
\begin{tabular}{lcccccccccccc}
\toprule
\multirow{2}{1cm}{{Models}} & \multicolumn{3}{c}{CNN/DM} & \multicolumn{3}{c}{SAMSum} & \multicolumn{3}{c}{SQuAD} & \multicolumn{3}{c}{XSum}\\
\cmidrule(lr){2-4}\cmidrule(lr){5-7}\cmidrule(lr){8-10}\cmidrule(l){11-13} 
& R-1 & R-2 & R-l & R-1 & R-2 & R-l & B-4 & R-L & MTR & R-1 & R-2 & R-l\\ 
\midrule
Ours (data+) & 47.18 & 23.51 & 43.78 & 53.49 & 28.60 & 49.79  & 27.80 & 54.86 & 28.14 & 48.29 & 25.28 & 40.27 \\
Ours (data-) & 48.54 & 24.40 & 45.35 & 54.27 & 29.19 & 50.57 & 27.79 & 54.95 & 28.32 & 49.02 & 25.37 & 40.52 \\
\bottomrule
\end{tabular}
}
\end{sc}
\end{small}
\end{center}
\vskip -0.1in
\end{table*}
\subsubsection{Different Options of Offline Dataset}
\label{sec:data+-}
In this part, we investigate the effect of using different static datasets for model optimization. Consider $K$ candidate targets obtained using diverse beam search or top-p sampling, etc. We sort them by metrics such as ROUGE and collect all sentences with the lowest metric among the K candidates as \textbf{\textsc{data-}} and the highest metric as \textbf{\textsc{data+}}. For our experiment, we follow BRIO~\cite{liu2022brio}, using diverse beam search to get the dataset, and more details can be referred to in Appendix~\ref{appendix:b}.

As shown in Tab.~\ref{cond2}, using \textsc{data-} as the dataset gives better results than \textsc{data+}. From an optimization perspective, we believe that \textsc{data-} is easier to sample useful signals, because the probability that it learns about how to further improve the sentence is higher.
\begin{table}[h]

\caption{Win rate of sampled text compared with greedy.}
\label{cond}
\begin{center}
\begin{small}
\begin{sc}
\resizebox{1\linewidth}{!}{
\begin{tabular}{lcccc}
\toprule
\multicolumn{1}{c}{{Models}} & \multicolumn{1}{c}{CNN/DM} & \multicolumn{1}{c}{SAMSum} & \multicolumn{1}{c}{SQuAD} & \multicolumn{1}{c}{XSum}\\
\midrule
Ours (data+) & 27 \% & 15\% & 13\% & 11\% \\
Ours (data-) & 32 \% & 19\% & 18\% & 16\% \\
\bottomrule
\end{tabular}
}
\end{sc}
\end{small}
\end{center}

\end{table}
To provide a numerical analysis, we calculate the proportion of sampled sentences that are better than the greedy decoding result (i.e. better than the current policy) in Tab.~\ref{cond}. We found that \textsc{data-} is more likely to sample better sentences for improving the current strategy. Even though we fix the data as input, the optimization is not only for these sentences. The mask information given by our environment each time is random and does not allow the model to see a complete and fixed sentence, which may represent more abstract semantics and prevent overfitting, as per Lemma~\ref{lemma:overfit}. Additionally, even though we only perform exploration on this data, the generalization ability of the neural model also facilitates the results on the test set.

For the optimization of text similarity towards ground truth, we use the aforementioned pre-decoded results as the static dataset following BRIO. In contrast, we can directly use ground truth as the dataset to in turn remove the pre-decoding cost in more general tasks. We test the results of optimizing factuality for summarization and click-through rate for textual advertisement. The results are provided in Appendix~\ref{appendix:other_task}.

\subsection{Sensitivity Analysis}
Tab.~\ref{sensitivity} shows how different numbers of samples impact model performance. We observe that when the same number of samples is used, we usually have better results compared with BRIO. This is probably due to the fact that we can more efficiently estimate the reward of each token, according to Proposition~\ref{prop:tk}. Furthermore, in contrast to BRIO, which requires more memory and FP cost to increase the number of samples, our sampling does not introduce additional memory and FP cost. So our method allows for a larger number of samples, e.g. 64, which results in further improved performance.
No significant benefit can be seen by continuing to increase the number of samples, e.g., 128.
\begin{table}[h]
\vspace{-4mm}
\caption{Sensitivity of \# sample on CNN/DM.}
\label{sensitivity}
\vskip 0in
\begin{center}
\begin{small}
\begin{sc}
\resizebox{0.90\linewidth}{!}{
\begin{tabular}{lcrccccc}
\toprule
\multirow{2}{1cm}{{Models}} & \multicolumn{5}{c}{CNN/DM} \\
\cmidrule(lr){2-6}  
& \# sample & FP & R-1 & R-2 & R-l \\ 
\midrule
\multirow{3}{1cm}{BRIO} & 2 & 2x & 46.02 & 22.21 & 42.71  \\
 & 4 & 4x & 46.10 & 22.31 & 42.85  \\
 & 16 & 16x & 47.83 & 23.75 & 44.65  \\
\midrule
\multirow{4}{1cm}{Ours} & 2 & 1x 
& 46.09&22.59&43.03\\
 & 4 & 1x 
& 47.05&23.33&43.87\\
 & 16 & 1x & 
48.31 & 23.91 & 45.15  \\
 & 64 & 1x & 48.54 & 24.40 & 45.35  \\
\bottomrule
\end{tabular}
}
\end{sc}
\end{small}
\end{center}
\vskip -0.2in
\end{table}

We give results of sensitivity for the choice of mask rate and the weight $\lambda$ in Appendix~\ref{appendix:sense}.
For mask rate, we use a default mask rate $0.4$ for the main experiment. As it performs well across various tasks. However, tuning the mask rate for the specific task can bring better results (Tab.~\ref{tab:maskrate}). For example, by setting the mask rate to $0.8$, we can get a better performance on SAMSum. For the weight $\lambda$,  one should tune this parameter as long as the weight is not too large or too small, the method can work well (Tab.~\ref{tab:lambda}). 
\section{Conclusion}
We propose semi-offline reinforcement learning, a novel paradigm that bridges the gap between online and offline RL, and provides a theoretical foundation for comparing different RL settings. Our semi-offline RL approach achieves a balance between effective exploration and minimum optimization cost. Extensive experiments show that our semi-offline RL approach is effective in various text generation tasks and datasets, and yields comparable or usually better performance compared to the state-of-the-art methods.

\section{Acknowledgement}
This work was supported by National Natural Science Foundation of China (NSFC Grant No. 62122089), Beijing Outstanding Young Scientist Program NO.BJJWZYJH012019100020098, and Intelligent Social Governance Platform, Major Innovation \& Planning Inter-disciplinary Platform for the ``Double-First Class" Initiative, Renmin University of China.


\bibliography{main}
\bibliographystyle{icml2023}


\clearpage
\appendix
\section{Proofs} 


\subsection{Proof of Proposition 1}
\label{appendix:a1}

\textbf{Proposition 1} (Time cost when fully observable)

Considering the fully observable scenario in which all information in the states is observed by the language model to get sampled tokens. Let us denote the minimum number of FPs required to sample $s_t$ as $C_t$ and its expectation as $\mathbb{E}(C_t)$. We have
\begin{equation}
    C_t=\sum_{t'=0}^{t-1}M_{t'},   \textnormal{     }
    \mathbb{E}(C_t)=tp_m
\end{equation}
where $M_{t'}\in \{0, 1\}$ denotes whether the next token $y^s_{t'}$ will be determined according to the agent's generation ($M_{t'}=1$) or the static dataset ($M_{t'}=0$). 
\begin{proof}
Our proof consists of three steps.

\underline{Step 1}. We first prove $C_t \geq \sum_{t'=0}^{t-1}M_{t'}$ by using mathematical induction.

Let us consider the scenario when $t=0$, and verify that $C_0=0$. As $s_0$ only contains the source $\textbf{x}$,  no FPs are needed, thus $C_0=0$ holds.
We can then easily see that for $t=1$,  $C_1 = M_0 \geq \sum_{t'=0}^{0} M_{t'}$.

Next, let us consider the scenario when $t>1$.
Assume that $C_{t-1} \geq \sum_{t'=0}^{t-2}M_{t'}$. There are two cases:
\begin{itemize}[nosep,leftmargin=1em,labelwidth=*,align=left]
    \item If $M_{t-1}=1$,  deriving $a_{t-1} \sim p(a_{t-1}|s_{t-1};\theta)$ needs 1 $FP$. In this case, $C_t = C_{t-1} + 1 = C_{t-1}+M_{t-1} \geq \sum_{t'=0}^{t-1}M_{t'}$, 
    \item If $M_{t-1}=0$,  $C_t$ can not be less than $C_{t-1}$,  $C_t \geq C_{t-1} = C_{t-1} + M_{t-1} \geq \sum_{t'=0}^{t-1}M_{t'}$.
\end{itemize}
In either case, $C_t \geq \sum_{t'=0}^{t-1}M_{t'}$ holds.

\underline{Step 2}. In the second step, we prove $C_t \leq \sum_{t'=0}^{t-1}M_{t'}$.

We can get all $M_t$ with 0 $FP$,  the tokens \{$y_{t'}^s: y_{t'}^s=y_t,  M_{t'}=0$\} are from the dataset and can also be derived with 0 FPs. What left is $\{y_{t'}^s: y_{t'}^s=\hat{y}_{t'},  M_{t'}=1\}$, 
whose size is $\sum_{t'=0}^{t-1}M_{t'}$,  so we can also get it with at most $\sum_{t'=0}^{t-1}M_{t'}$ FPs. Thus, $C_t \leq \sum_{t'=0}^{t-1}M_{t'}$.

\underline{Step 3}. Combing 1 and 2,  we have $C_t = \sum_{t'=0}^{t-1}M_{t'}$. Accordingly, $\mathbb{E}(C_t)=\mathbb{E}(\sum_{t'=0}^{t-1}M_{t'})=\sum_{t'=0}^{t-1}\mathbb{E}(M_{t'})=\sum_{t'=0}^{t-1}p_m = tp_m$.
\end{proof}

\subsection{Proof of Proposition 2}

\textbf{Proposition 2} (Information loss with minimum time cost)
\label{appendix:a2}

If $s_{T}$ can always be sampled within 1FP for $\forall p_m\in[0,1]$, then $o_t$ must \textbf{not} contain any exact information about sampled tokens $\hat{y}_{t'}$,
for $\forall t' \in [0, t-1]$ and $\forall t \in [1, T)$.
%
\begin{proof} 
    We prove that if $o_t$ contains information about a sampled token, then we need at least 2FP to compute $s_{T}$.
    
    \begin{itemize}[nosep,leftmargin=1em,labelwidth=*,align=left]
        \item For $t\in [1, T)$, 
    suppose that $o_t$ contains the exact information of some $\hat{y}_{t'}$.
    Then,  we need at least 1FP to derive $o_t$ (in order to get $\hat{y}_{t'}$).
        \item At time $t$,  with probability $p_m>0$, one needs to compute $a_{t}=\hat{y}_{t}$ given $o_t$. This requires an additional FP.
        \item Thus, with probability $p_m>0$, we need at least $\ 2\text{FP}$=$1\text{FP}$ (computing $\hat{y}_{t'}$)+$1\text{FP}$ (computing $a_{t}$) to derive $s_{t+1}$. Considering that $s_{t+1}$ is a subsequence of $s_{T}$, we also need at least 2FP to derive $s_{T}$.
    \end{itemize}
    
    
    
\end{proof}

\subsection{Proof of Theorem 1}
\label{appendix:a3}
\textbf{Theorem 1} (optimality of masked observation). $o^M_t$ in Def.~\ref{def:4} is $o^*_t$ in Def.~\ref{def:3}.

\begin{proof}
To prove Theorem 1, we prove $o^M_t$ in Def.~\ref{def:4} satisfies DP1-DP3 in Def.~\ref{def:3} as follows.

\underline{DP1}. Here we prove that we can compute $a_t$=$\hat{y}_{t}$ within 1FP.\looseness=-1

To get $a_t$=$\hat{y}_{t}$, we need to compute $p(\hat{y}_{t}|o^M_t;\theta)$ and sample $a_t$ from it:
$a_t$=$\hat{y}_{t}\sim  p(\hat{y}_{t}|o^M_t;\theta)$.
Please note that $o^M_t$ does not contain any information about sampled tokens (does not violate Proposition.~\ref{prop:2}) and can be derived with 0FP. Thus, $p(\hat{y}_{t}|o^M_t;\theta)$ can be computed by using 1FP. No matter how many actions $a_t$ we wish to sample from 
$p(\hat{y}_{t}|o^M_t;\theta)$, it can all be done without further FPs as long as $p(\hat{y}_{t}|o^M_t;\theta)$ is determined. Thus, we only need 1FP to get $a_t=\hat{y}_t$.

\underline{DP2}. To prove that we achieve the minimum asymmetric bias, we need to prove $p(s|o_{t}^{max})=p(s|o_t^M)$ according to Lemma~\ref{lemma:asymmetric bias}.
Since the state $s$ is a trajectory of tokens, we have the following two equations:
\begin{equation}
p(s|o_t^M)=\prod_{t'} p(y^s_{t'}|o^M_{t})\\
\label{proof:a.3.2.1.1}
\end{equation}
\begin{equation}
p(s|o_{t}^{max})=\prod_{t'} p(y^s_{t'}|o^{max}_{t})\\
\label{proof:a.3.2.1.2}
\end{equation}
Then, we consider if the $t'$-th token is a generated token or a token from the static data, in order to compute $p(y^s_{t'}|o^M_{t})$ and $p(y^s_{t'}|o^{max}_{t})$. When $y^M_{t'}=\text{[M]}$ or $M_t=1$, $y_{t'}^s$ is derived by sampling from $p(y^s_{t'}|x, y_{<t' \verb|\| m})$. When $y^M_{t'}\neq \text{[M]}$ or $M_t=0$, $y_{t'}^s$ is the same as the input $y_{t'}^M$, i.e., the same as the static token $y_{t'}$ from dataset. Thus, we have:


\[
\begin{array}{l}
p(y^s_{t'}|o^M_{t})=\\
\quad \left\{
\begin{array}{rcl}
p(y^s_{t’}|o^M_{t’-1})=p(y^s_{t'}|x, y_{<t' \verb|\| m}) & & \text{if } y^M_{t'}=\text{[M]}\\
p(y_{t'}^s=y_{t'}^M)=1 & & \text{if } y^M_{t'}\neq \text{[M]}\\
0 & & \text{otherwise}
\end{array} \right.
\end{array}
\label{proof:a.3.2.1}
\]



\begin{equation}
p(y^s_{t'}|o^{max}_{t})=\left\{
\begin{array}{rcl}
p(y^s_{t'}|o^{max}_{t'-1})=p(y^s_{t'}|x, y_{<t' \verb|\| m}) & M_t=1\\
p(y_{t'}^s=y_{t'}^M)=1 &  M_t=0\\
0 & \text{otherwise},  \\
\end{array} \right.
\label{proof:a.3.2.2}
\end{equation}

According to Eq.~\eqref{proof:a.3.2.1} and Eq.~\eqref{proof:a.3.2.2}, we have $p(y^s_{t’}|o^M_{t})=p(y^s_{t'}|o^{max}_{t})$. Together With Eq.~\eqref{proof:a.3.2.1.1} and Eq.~\eqref{proof:a.3.2.1.2}, we have $p(s|o_{t}^{max})=p(s|o_t^M)$.

\underline{DP3}. According to Lemma~\ref{lemma:overfit}, we can prove that we achieve the minimum overfitting error bound by proving that the observation space of $o^M_t$ (i.e., $|O_t^M|$) is minimum for all observations that satisfy DP1 and DP2.
We prove this by illustrating that removing features in $o^M_t$ will result in the violation of DP2.


First, the size of the observation space is $|O^M_t|=(|\mathcal{V}|+1)^{t+1}$. Take $t=0$ for an instance. We have $o_t^M=(x, M_0, y^{M}_0)$, and $|O^M_t|=|\mathcal{V}|+1$, i.e., the observation space includes tokens from the vocabulary and a token [M]. Then we remove information from $o_t^M=(x, M_0, y^{M}_0)$ and prove that DP2 will be violated.

1. If $y^{M}_0$ is removed from the observation, then the new observation becomes $\phi'(o_t^{max})=(x, M_0)$. In this case,

\begin{equation}
p(y_{t'}^s|\phi'(o_t^{max}))=\left\{
\begin{array}{rcl}
p(y_{t'}^s|x, M_0) & & y^M_{t'}=\text{[M]}\\
\frac{1}{|\mathcal{V}|} & &  y^M_{t'}\neq\text{[M]}\\
\end{array} \right.
\label{eq:DP3_1}
\end{equation}

We have $p(y_{t'}^s|\phi'(o_t^{max})) \neq p(y_{t'}^s|o_t^{max})$ according to Eqs.~(\ref{proof:a.3.2.2}) and (\ref{eq:DP3_1}), thus DP2 is violated.

2. If $M_0$ is removed from the observation, then the new observation becomes  $\phi^{''}(o_t^{max})=(x, y^M_0)$. In this case,

\[
\begin{array}{l}
p(y _{t'}^s|\phi^{''}(o_t^{max}))=\\
\quad \left\{
\begin{array}{rcl}
p_m p(y_{t'}^s|x, \tilde{y}_{<t})+(1-p_m) & \text{if } y^M_{t'}=y^s_{t'}\\
p_m p(y_{t'}^s|x, \tilde{y}_{<t}) & \text{if } y^M_{t'}\neq y^s_{t'}\\
\end{array} \right.
\end{array}
\label{eq:DP3_2}
\]

We have $p(y_{t'}^s|\phi''(o_t^{max})) \neq p(y_{t'}^s|o_t^{max})$ according to Eqs.~(\ref{proof:a.3.2.2}) and (\ref{eq:DP3_2}), thus DP2 is violated.

In summary, $|O^M_t|$ is the minimum one among all that satisfy DP1 and DP2, since removing any information in $o_t^{M}$ will result in the violation of DP2.

\end{proof}


\subsection{Proof of Proposition~\ref{prop:tk}}
\label{appendix:a4}
\textbf{Proposition 3} (Token-level reward assignment)
$\mathcal{L}_{RL}$ in Eq.~\ref{eq:rl_f} can be decomposed into token-wise loss $\mathcal{L}_{t}^i$ of the $i$-th token in the vocabulary at time step $t$: 
\begin{equation*}
\begin{array}{lll}
\mathcal{L}_{RL} = \sum_{t} \sum_{i=1}^{|\mathcal{V}|}\mathcal{L}_{t}^i\\
\mathcal{L}_{t}^i= -\frac{C_t^i}{K}\ \log \ 
p(\mathcal{V}_i|o_t^M) (\frac{\sum_{k=1, \mathcal{V}_i=a_t^k}^K  \mathcal{R}(Y^k)}{C_t^i}-b) \\
b=\frac{\sum_{k=1}^K\mathcal{R}(Y^k)}{K} \\
\end{array}
\end{equation*}

\begin{proof}
We first write down Eq.~(\ref{eq:rl_f}):
\begin{equation}
    \begin{array}{cc}
\mathcal{L}_{RL}=\frac{1}{K}\sum_{k=1}^K - (\mathcal{R}(Y^k)-b) \sum_t \log \ p(a^k_t|o_t^M) \\
=\frac{1}{K}\sum_{k=1}^K\sum_t - (\mathcal{R}(Y^k)-b)  \log \ p(a^k_t|o_t^M) \\
=\frac{1}{K}\sum_t\sum_{k=1}^K - (\mathcal{R}(Y^k)-b)  \log \ p(a^k_t|o_t^M) \\
=\sum_t\frac{1}{K}\sum_{k=1}^K - (\mathcal{R}(Y^k)-b)  \log \ p(a^k_t|o_t^M) \\
=\sum_t \mathcal{L}_{t}
    \end{array}
\end{equation}
where $\mathcal{L}_{t}$ is the loss for some specific time step $t$:
\begin{equation*}
 \begin{array}{cc}
\mathcal{L}_{t}=\frac{1}{K}\sum_{k=1}^K - 
(\mathcal{R}(Y^k)-b)  \log \ p(a^k_t|o_t^M) \\
\end{array}
\end{equation*}

Then we add the enumeration of the actions. $\mathcal{V}_i$ denotes the $i$-th action in the action space (vocabulary space). When sampling multiple $Y$, for time step $t$,  one specific $\mathcal{V}_i$ may appear in multiple $Y$.
\begin{equation*}
\begin{array}{cc}
\mathcal{L}_{t}=\frac{1}{K}\sum_{k=1}^K \sum_{i=1, \mathcal{V}_i=a_t^k}^{|\mathcal{V}|} - (\mathcal{R}(Y^k)-b) \log \ p(a^k_t|o_t^M) \\
=\frac{1}{K}\sum_{i=1}^{|\mathcal{V}|}\sum_{k=1, \mathcal{V}_i=a_t^k}^K - (\mathcal{R}(Y^k)-b)  \log \ p(a^k_t|o_t^M) \\
=\sum_{i=1}^{|\mathcal{V}|}\frac{1}{K} \sum_{k=1, \mathcal{V}_i=a_t^k}^K - (\mathcal{R}(Y^k)-b)  \log \ p(a^k_t|o_t^M) \\
=\sum_{i=1}^{|\mathcal{V}|}\mathcal{L}_{t}^i\\
\end{array}
\end{equation*}
where $\mathcal{L}_{t}^i$ satisfies
\begin{equation*}
    \begin{array}{cc}
\mathcal{L}_{t}^i=\frac{1}{K}\sum_{k=1, \mathcal{V}_i=a_t^k}^K - (\mathcal{R}(Y^k)-b)  \log \ p(a^k_t|o_t^M) \\
=\frac{1}{K}\sum_{k=1, \mathcal{V}_i=a_t^k}^K - (\mathcal{R}(Y^k)-b)  \log \ p(\mathcal{V}_i|o_t^M) \\
    \end{array}
\end{equation*}

As one $\mathcal{V}_i$ may appear in multiple $Y$. We can assume in these samples,  what we do is fix $\mathcal{V}_i$ at time step $t$,  and do sampling at other positions. We regard the sample results of other positions as different contexts for $\mathcal{V}_i$ at time step $t$. Then we can compute the expected reward of $\mathcal{V}_i$ using these samples.
\begin{equation*}
    \begin{array}{cc}
\mathcal{L}_{t}^i= -\log \ p(\mathcal{V}_i|o_t^M) \frac{1}{K}\sum_{k=1, \mathcal{V}_i=a_t^k}^K  (\mathcal{R}(Y^k)-b) \\
    \end{array}
\end{equation*}

From Eq.~\ref{eq:rl_f},  we have
\begin{equation*}
b=\frac{\sum_{k=1}^K\mathcal{R}(Y^k)}{K}
\end{equation*}

Let\ $C_t^i=\sum_{k=1}^K\mathbb{I}(\mathcal{V}_i=a_t^k)$, then we get the formulation of Eq.~\eqref{prop:4}:
\\
\begin{equation*}
    \begin{array}{cc}
\mathcal{L}_{t}^i= -\frac{C_t^i}{K}\ \log \ 
p(\mathcal{V}_i|o_t^M) (\frac{\sum_{k=1, \mathcal{V}_i=a_t^k}^K  \mathcal{R}(Y^k)}{C_t^i}-b) \\
    \end{array}
\end{equation*}

We compute $\frac{\sum_{k=1, \mathcal{V}_i=a_t^k}^K  \mathcal{R}(Y^k)}{C_t^i}-\frac{\sum_{k=1}^K\mathcal{R}(Y^k)}{K}$ to estimate $\mathbb{E}_{Y \sim p(Y|o=o^M_t,a=\mathcal{V}_i)}\mathcal{R}(Y)-\mathbb{E}_{Y \sim p(Y|o=o_t^M)}{\mathcal{R}(Y)}$:

If\ $C_t^i \to \infty$ and $\ K \to \infty$, we have\\
\begin{equation*}
\lim_{{K \to \infty}} \frac{\sum_{k=1}^K\mathcal{R}(Y^k)}{K}=\mathbb{E}_{Y \sim p(Y|o=o_t^M)}{\mathcal{R}(Y)}
\end{equation*}
\begin{equation*}
\lim_{{K \to \infty, C_t^i \to \infty}} \frac{\sum_{k=1, \mathcal{V}_i=a_t^k}^K  \mathcal{R}(Y^k)}{C_t^i}=\mathbb{E}_{Y \sim p(Y|o=o^M_t,a=\mathcal{V}_i)}\mathcal{R}(Y)
\end{equation*}

The term $\mathbb{E}_{Y \sim p(Y|o=o^M_t,a=\mathcal{V}_i)}\mathcal{R}(Y)
-\mathbb{E}_{Y \sim p(Y|o=o_t^M)}{\mathcal{R}(Y)}$ is the expected advantage of generating $\mathcal{V}_i$ under observation $o_t^M$. 
\end{proof}
\section{Implementation Details}
\label{appendix:b}
\begin{table}[h]
\vspace{-4mm}
\caption{Implementation details. For CNN/DM, SAMSum, and XSum the ROUGE score is the average of ROUGE-1,  ROUGE-2 and ROUGE-L. For Squad. the Reward is the average of BLEU-4 and ROUGE-L.}
\label{detail}
\vspace{-4mm}
\begin{center}
\begin{small}
\begin{sc}
\resizebox{1\linewidth}{!}{
\begin{tabular}{lcccc}
\toprule
\multicolumn{1}{c}{{-}} & \multicolumn{1}{c}{CNN/DM} & \multicolumn{1}{c}{SAMSum} & \multicolumn{1}{c}{SQuAD} & \multicolumn{1}{c}{XSum}\\
\midrule
Batch size & 16 & 16 & 16 & 16 \\
Learning rate & 1e-6 & 1e-6 & 3e-6 & 2e-6 \\
$\lambda$(weight of $\mathcal{L}_{RL}$) & 20 & 5 & 4 & 2 \\
\# sample & 64 & 64 & 16 & 64 \\
$p_m$(mask rate) & 0.4 & 0.4 & 0.4 & 0.4 \\
Reward & ROUGE & ROUGE & BLEU-4 + ROUGE & ROUGE\\
\bottomrule
\end{tabular}
}
\end{sc}
\end{small}
\end{center}
\vskip -0.1in
\end{table}

Here,  we introduce how we get the trajectories for compared methods and our method.

\textbf{Offline} We evaluate two offline methods: GOLD,  BRIO. For GOLD,  we follow their original setting ~\cite{pang2020text} and use the ground truth labels as the samples. For BRIO,  we follow its settings ~\cite{liu2022brio} to obtain the samples. In particular,  given a base model trained with only supervised learning (more details in Sec.~\ref{sec:cmp},  ``Base models''),  we obtain samples from the base model by using Diverse Beam Search ~\cite{Vijayakumar2016DiverseBS}. This produces $K$ (=16) diverse output sentences for each input. 

\textbf{Semi-offline} ground-truth labels or pre-generated results can both be used as our dataset. In our experiment,  we use the same pre-generated results with BRIO. BRIO uses all outputs as the samples for offline learning,  while we only select one as the initial exploration point for semi-offline training. Based on the experiment results in Sec.~\ref{sec:data+-},  we decide to use the sample with the lowest reward (i.e.,  DATA-) as our static data,  since it leads to a higher probability of learning about improvement directions.
Then we mask the static data as the input of our model. Our model predicts the action distribution of each mask position,  and samples these distributions simultaneously to generate the final trajectories. For example,  given an offline data point consisting of four tokens: A,  B,  C,  and D,  we randomly mask it and obtain for example A [M1] B [M2]. The model then calculates the action distribution for [M1] and [M2]. We sample tokens from the two distributions independently multiple times,  resulting in multiple trajectories. 

\textbf{Online} In the online approach,  the trajectory is generated via real-time sequential decoding and sampling. We employ the same decoding parameters as those utilized during evaluation. We follow previous work to set these hyper-parameters~\cite{gliwa-etal-2019-samsum,liu2022brio,ushio-etal-2022-generative}. For example, we set a beam size of 4,  minimum length of 56,  maximum length of 142,  and other relevant parameters for the CNN/DM using BART.

\section{Datasets Statistics}
\label{appendix:c}
\begin{table}[h]
\vspace{-5mm}
\caption{Statistical information on the datasets.}
\label{stat}
\vspace{-3mm}
\begin{center}
\begin{small}
\begin{sc}
\resizebox{1\linewidth}{!}{
\begin{tabular}{lcccc}
\toprule
\multicolumn{1}{c}{{-}} & \multicolumn{1}{c}{CNN/DM} & \multicolumn{1}{c}{SAMSum} & \multicolumn{1}{c}{SQuAD} & \multicolumn{1}{c}{XSum}\\
\midrule
\# Train & 287, 113 & 14, 732  & 75, 722 & 204, 045 \\
\# Dev & 13, 368 & 818 & 10, 570 & 11, 332  \\
\# Test & 11, 490 & 819 & 11, 877 & 11, 334 \\
$|Source|$ & 781 & 124 & 148 & 431 \\
$|Target|$ & 56 & 23 & 11 & 23
 \\
\bottomrule
\end{tabular}
}
\end{sc}
\end{small}
\end{center}
\vspace{-7mm}
\end{table}

\section{Deriving AVG from BRIO}
\label{appendix:d}
We introduce how to get the RL loss of AVG from the contrastive loss of BRIO~\cite{liu2022brio}.
We first give the original loss function in BRIO.
\begin{equation*}
    \begin{array}{cc}
\mathcal{L}_{ctr}=\sum_i\sum_{j>i}max(0,  f(S_j)-f(S_i)+\lambda_{ij})\\
f(S)=\frac{\sum_{t=1}^l \log \ p_{g_\theta}(s_t|D, S<t;\theta)}{|s|^{\alpha}}\\
    \end{array}
\end{equation*}

 There are $N$ samples,  $S_i$ is the $i$-th sample,  $f(S)$ is the normalized sum of the loglikelihood of tokens in $S$,  $|S|$ denote the length of $S$,  $\alpha$ and $\lambda$ are two hyperparameters. the sentences are sorted by some quality metric $M$,  and $M_i > M_j$.

 Let's consider the loss $\mathcal{L}_{ij}$ for each pair of samples $(i, j)$. The function $max(0,  .)$,  together with the  sorted results,  gives a desired ordering condition: when $ f(S_j)-f(S_i)+\lambda_{ij} > 0$,  $M_j > M_i$ should hold.  It means when the order of the sum of loglikelihood $<f(S_j),  f(S_i)>$ disobeys the order of quality $<M_i, M_j>$,  we should rerank $S_i$ and $S_j$.
 Then we have:
\begin{equation}
    \begin{array}{cc}
\mathcal{L}_{ij}=\mathbb{I}(M_i>M_j\ \&\  f_i<f_j + \lambda_{ij})(\frac{\log  P_j}{|s_j|^\alpha}-\frac{\log  P_i}{|s_i|^\alpha})\\
    \end{array}
\label{eq:avgbrio}
\end{equation}
where $\mathcal{L}_{ctr}=\sum_i\sum_{j>i}\mathcal{L}_{ij}$. $f_i$ represents $f(S_i)$ and $\log  P$ represents $\log  \ p_{g_\theta}(s_t|D, S<t;\theta)$ for simplification. 

We can change this desired ordering condition so that we can derive the formulation of RL loss of AVG: 

1. To remove $f_i<f_j + \lambda_{ij}$ in $\mathcal{L}_{ij}$,  we consider that no matter whether the ordering of $f$ is correct,  if the model has a stochastic policy,  we should keep increasing the probability of the best action to get a better return. 

2. $\mathbb{I}(M_i>M_j)$ gives a discrete signal,  which we can replace with a continuous signal $M_i-M_j$.

Then, 
\begin{equation*}
    \begin{array}{cc}
\mathbb{I}(M_i>M_j \ \&\  f_i<f_j+\lambda_{ij})\approx M_i-M_j\\
    \end{array}
\end{equation*}

We replace $\mathbb{I}(M_i>M_j \ \&\  f_i<f_j+\lambda_{ij})$ with the new term $M_i-M_j$ in Eq.\eqref{eq:avgbrio}:
\begin{equation*}
    \begin{array}{ll}
\mathcal{L}=\sum_{i, j}\mathcal{L}_{ij}\\
=\sum_{i, j}(M_i-M_j)(\frac{\log P_j}{|s_j|^\alpha}-\frac{\log  P_i}{|s_i|^\alpha})\\
=\sum_{i, j}(\frac{\log P_j}{|s_j|^\alpha})(M_i-M_j)+\sum_{i, j}-(\frac{\log P_i}{|s_i|^\alpha})(M_i-M_j)\\
=\sum_{i, j}-(\frac{\log P_j}{|s_j|^\alpha})(M_j-M_i)+\sum_{i, j}-(\frac{\log P_i}{|s_i|^\alpha})(M_i-M_j)\\
=2\sum_{i, j}-\frac{\log P_j}{|s_j|^\alpha}(M_j-M_i)\\
=2\sum_i\sum_j-\frac{\log P_j}{|s_j|^\alpha}(M_j-M_i)\\
=2\sum_j-\frac{\log P_j}{|s_j|^\alpha}(\sum_iM_j-\sum_iM_i)\\
=2\sum_j-\frac{\log P_j}{|s_j|^\alpha}(NM_j-\sum_iM_i)\\
=2N\sum_j-\frac{\log P_j}{|s_j|^\alpha}(M_j-\frac{\sum_iM_i}{N})\\
    \end{array}
\end{equation*}


For the $j$-th sample,  $\mathcal{L}_j =-\frac{2N}{|s_j|^\alpha} \log P_j(M_j-\frac{\sum_iM_i}{N})$.\\

$\frac{\sum_iM_i}{N}$ can be regarded as the baseline averaging the reward of N samples. It is in a formulation of REINFORCE with baseline and is the same as the loss we use in Eq.~\ref{eq:rl_f} if we ignore the coefficient. 

For our compared method AVG in Sec.~\ref{sec:exp},  we use this training loss for optimization.


\section{Discussion of Limitations}

\textbf{Parallel prediction of future tokens}:
One potential limitation of our method is that the parallel prediction of future tokens may result in a lack of fluency in the sentences. We give a case of the parallel prediction in Tab.~\ref{case1}. The repetition ``will will'' and ``Friday Friday'' happen when many [M] tokens are connected together. Long masked sequence is difficult for a parallel decoder,  since it needs to generate all masked tokens together in one forward propagation. 
However,  we believe that using a multi-layer transformer model as the base model can alleviate this issue. 
The generative model can be regarded as a stack of two pre-trained $K$-layer transformer models, resulting in a 2$K$-layer model.
The first $K$-layers of the model make their own predictions for generation,  while the last $K$-layers take into account predictions from the previous time step,  leading to more informed predictions. By only estimating the action distribution information from the previous time step,  the model effectively models the unknown state through estimation in the intermediate layers of the transformer, similar to a belief function in the POMDP theory. 

In terms of FP,  assume the unit changes from the whole model to one layer for a $K$-layer transformer model. Therefore,  1 model-level FP is equivalent to $K$ layer-level FPs. While according to Theorem~\ref{theorem:mask},  the model cannot access the sequential self-generated information under the 1-FP setting,  at the layer level,  the higher layer can access some of the sequential information from the lower layer during the $K$ FPs.

For the majority of conditional generation tasks,  our method can be applied and optimize the reward. We also acknowledge the limitation of our method in employing sampling-based techniques for generating coherent long-form text. While our methods may be highly effective in optimizing local aspects of a particular metric,  we concede that the optimization of long dependencies remains a challenging task due to the potential for incoherent trajectories.

\begin{table}[h]
\vspace{-4mm}
\caption{Case 1. A  very long masked sequence may result in repetition.}
\label{case1}
\vspace{-4mm}
\begin{center}
\resizebox{1\linewidth}{!}{
\begin{tabular}{lp{0.4\textwidth}}
\toprule
Original data & Martha will be in Cracow on Thursday morning and will stay there until Friday afternoon.\\
Masked data & Martha will be in Cracow on [M] [M] [M] [M] [M] [M] [M] [M] [M] \\
Prediction & Martha will be in Cracow on Thursday morning and will will be stay Friday Friday \\
\bottomrule
\end{tabular}
}
\end{center}
\vskip -0.1in
\end{table}

\begin{table}[h]
\vspace{-4mm}
\caption{Case 2. Tokens generated at time 
 may be inconsistent with static data given after 
,  due to the unidirectional attention in the decoder.}
\label{case2}
\vspace{-4mm}
\begin{center}

\resizebox{1\linewidth}{!}{
\begin{tabular}{lp{0.4\textwidth}}
\toprule
Original data & Hee wants to move in with a man she knows for 2 months to another state. Jane strongly disagrees.\\

Masked data & [M] wants to move in with a [M] she [M] [M] 2 [M] [M] another state. Jane [M] disagrees [M] \\

Prediction & Hee wants to move in with a guy she knows for 2 months. another state. Jane is disagrees. \\
\bottomrule
\end{tabular}
}
\end{center}
\vspace{-2mm}
\end{table}

\begin{figure}
  \includegraphics[width=0.46\textwidth]{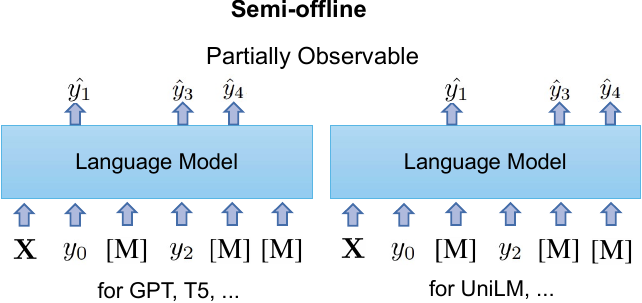}
  \vspace{-1mm}
  \caption{Our method can be applied to different architectures built on Transformer.
  }
  \vspace{-3mm}
  \label{fig:cmp_rl_a}
\end{figure}

\textbf{Relevance of tokens from dataset and model prediction}: Another possible disadvantage of our approach is that if data replacement is performed after a period of model generation,  the current data may not be relevant to the previous model generation. We show a case for this irrelevance in Tab.~\ref{case2}. the end of the predicted sentence ``. another state. Jane is disagrees.'' is not fluent. Such a broken sentence is generated because the model fail to foresee ``another'' and ``disagrees'' when generating the token before them. Please note that this fluency issue only happens during training,  and the model will learn to avoid such broken generation based on the low reward. During testing,  we use sequential decoding to avoid this issue. 
In light of this mismatch, our approach can be seen as optimizing each fragment of the target. However,  as each fragment is predicted by the token from the same data,  the correlation between each fragment is partially preserved. We have experimented with the general case,  where $p_m$ randomly determines the sequence of masks,  and have achieved good results with the current experimental settings. 

We suggest further experimentation with mask replacement,  such as masking only the end of sentences or specific parts for better results,  considering that inter-sentence associations are weaker than intra-sentence associations. For practical applications,  such as advertisements generation,  the adjective or numerical parts of the sentences could be masked and optimized to generate more attractive or factual descriptions. 
Meanwhile,  the generative models considered in our experiments are not bidirectional,  and the optimization method does not affect the model structure. 
In this sense, the use of bidirectional models can be considered,  but would require changes to the model structure and inference method.
Or one can use another auxiliary bidirectional model as the behaviour policy to get multiple samples, but it increases the FPs to NK for computing the logits of these multiple samples like BRIO.
In our method, the generative models we consider are not bidirectional, and the method doesn't affect the model structure or require more FPs.

\textbf{Limited exploring space}
Semi-offline methods indeed result in a smaller exploration space,  which we mentioned at the end of Sec.~\ref{sec:form} with a quantitative comparison: ``the space to be explored for semi-offline methods ($|\mathcal{V}|^{Tp_m}$) is exponentially smaller than that of online methods ($|\mathcal{V}|^{T}$)''.  Our method may lead to potential suboptimal solutions due to the incapability to thoroughly investigate the entire space.

We wish to highlight here that this limitation may be largely alleviated by the generalization ability of deep models,  and that the semi-offline setting with a small exploration space may yield more benefits than issues. This is discussed in Sec.~\ref{sec:form}: ``(the smaller exploration space makes) it easier for the language model to understand the reward gain brought by different choices. Even though the exploration space is limited,  it is possible that the knowledge explored in the vicinity of specific output text can be generalized to other output text considering the generalization ability of neural networks. This is verified by our experiments,  which show that semi-offline usually performs equally well or better with much less time cost compared with existing online or offline methods (Sec.~\ref{sec:exp}).''

The good performance of BRIO also demonstrates that the generalization ability of neural networks may be leveraged to avoid exploring every point. 

\textbf{Negative societal impact} While our text generation method has exhibited promising results in text generation,  it is important to consider its potential negative impact on society. The generated text could be utilized for spreading misinformation,  reinforcing negative biases,  or serving other malicious purposes. Given the risk of misinformation,  it is crucial to establish safeguards such as fact-checking mechanisms. In addition,  we recommend incorporating fairness principles into the reward function to mitigate potential biases in the generated content. Furthermore,  addressing nefarious use cases requires the implementation of monitoring systems to prevent misuse and protect public discourse.

\textbf{Hard to optimize length} Furthermore,  we have identified that our approach may not be suitable for length optimization. The generation is performed by providing a pre-defined generation length based on the offline data. It is not clear whether our method could effectively make a target sequence longer or shorter,  which we are interested in investigating in the future.

\section{Sensitivity Analysis} 
\label{appendix:sense}
\begin{table}[h]
\vspace{-4mm}
\caption{Sensitivity of mask rate.}
\label{tab:maskrate}
\vspace{1mm}
\begin{center}
\begin{sc}
\resizebox{0.8\linewidth}{!}{
\begin{tabular}{lcccc}
\toprule
mask rate & R-1       & R-2       & R-L       \\ \midrule
OURS ($p_m$=0.1) & 53.75     & 28.92     & 49.54     \\ 
OURS ($p_m$=0.2) & 53.96     & 28.82     & 49.89     \\ 
OURS ($p_m$=0.3) & 53.93     & 29.10     & 50.10     \\ 
OURS ($p_m$=0.4) & 54.27     & 29.19     & 50.57     \\ 
OURS ($p_m$=0.5) & 54.21     & 29.10     & 50.45     \\ 
OURS ($p_m$=0.6) & 54.46     & 29.11     & 50.43     \\ 
OURS ($p_m$=0.7) & 54.64 & \textbf{29.69} & 50.89 \\
OURS ($p_m$=0.8) & \textbf{54.70}     & 29.25     & \textbf{51.00}     \\ 
OURS ($p_m$=0.9) & 54.19     & 29.40     & 50.61     \\ 
OURS ($p_m$=1) = NAT   & 53.55     & 28.87     & 49.54   \\  
\bottomrule
\end{tabular}
}
\end{sc}
\end{center}
\vskip -0.1in
\vspace{-1mm}
\end{table}

Tab.~\ref{tab:maskrate} shows tuning the mask rate can bring better results on SAMSum than the default mask rate (0.4). 
Note that when $p_m=1$,  it becomes an online RL method with Non-autoregressive Transformer (NAT) that never uses static data during training,  where $p_m$ is the probability that we use the generated token (to perform exploration) instead of leveraging the static data point (to find a good initial point). The results also show the importance of using semi-offline training ($0<p_m<1$) instead of the online one ($p_m=1$). We can see that the semi-offline setting outperforms the online one (NAT) with a wide variety of $p_m$. Moreover,  the performance first increases with increasing $p_m$ and then decreases when $p_m$ becomes too large,  which further demonstrates the necessity to balance exploration and the effective leverage of offline static dataset with a semi-offline setting.
\begin{table}[h]
\caption{Sensitivity of the interpolation weight on SAMSum and SQuAD. Here we report the R-L scores. Results on other criteria are similar.}
\label{tab:lambda}
\begin{center}
\begin{sc}
\resizebox{0.8\linewidth}{!}{
\begin{tabular}{lcccc}
\toprule
\multirow{2}{1cm}{{Weight}} & \multicolumn{2}{c}{SAMSum} & \multicolumn{2}{c}{SQuAD}\\
\cmidrule(lr){2-3}\cmidrule(lr){4-5}
& OURS & AVG & OURS & AVG \\ 

\midrule
0 (Base) & 48.98 & 48.98 & 54.30 & 54.30 \\ 
0.1 & 48.76 & 49.04 & 54.50 & 54.26 \\ 
1 & 49.72 & 49.18 & 54.75 & 54.68 \\
2 & 50.35 & 49.58 & 54.87 & \textbf{54.79} \\ 
3 & 50.33 & \textbf{49.65} & 54.84 & 54.73 \\
4 & 50.45 & 49.25 & \textbf{54.95} & 54.65 \\ 
5 & \textbf{50.57} & 49.03 & 54.92 & 54.71 \\ 
10 & 50.31 & 49.06 & 54.54 & 54.36 \\ 
\bottomrule
\end{tabular}
}
\end{sc}
\end{center}
\vskip -0.1in
\end{table}
For the weight $\lambda$,  we show how our method and the most competitive baseline AVG perform with varying interpolation weight (Tab.~\ref{tab:lambda}). As shown in the table,  our method is better than AVG for most of the weight values. Moreover,  our method yields comparable results as reported in the paper for a wide range of weight values (1-5). For both AVG and our method,  the performance first increases with increasing interpolation weight,  and then drops after the weight becomes too large. This trend verifies the necessity to balance the MLE and RL loss and shows a clear pattern that helps better understand the relationship between performance and weight.

\section{Experiments on Other Rewards and Tasks}
\label{appendix:other_task}
To show our method can be applied to optimize other rewards in addition to text similarity towards ground truth (e.g.,  ROUGE and BLEU),  we experiment with two other rewards. 
\begin{table}[h]
\vspace{-4mm}
\caption{Optimizing Factuality on CNN/DM.  $*$ indicates the metrics directly optimized during training.}
\label{tab:fact}
\begin{center}
\begin{small}
\resizebox{0.8\linewidth}{!}{
\begin{tabular}{lcccc}
\toprule
&  R-1 & R-2 & R-l & Fact*\\
\midrule
BASE & 45.10 & 21.76 & 41.86	& 15.57  \\
BRIO & 44.23	 & 21.18 & 
 40.92	& 17.54  \\
AVG & 44.23	 & 21.24 & 40.93	& 17.68  \\
OURS & 44.60	 & 21.66 & 41.91	& 18.30  \\
\bottomrule
\end{tabular}
}
\end{small}
\end{center}
\vspace{-9mm}
\end{table}

\begin{table}[h]
\caption{Optimizing CTR for advertisement generation.  $*$ indicates the metrics directly optimized during training.}
\label{tab:ctr}
\vspace{-3mm}
\begin{center}
\begin{small}
\resizebox{1\linewidth}{!}{
\begin{tabular}{lccccc}
\toprule
& R-1   & R-2   & R-L   & Distinct & CTR* \\ 
\midrule
Base              & 34.22 & 11.44 & \textbf{28.21} & 0.578   & 0.1998                   \\ 
OURS              & \textbf{34.51} & \textbf{11.45} & \textbf{28.21} & \textbf{0.590} & \textbf{0.2086}          \\ 

\bottomrule
\end{tabular}
}
\end{small}
\end{center}
\vskip -0.1in
\end{table}
1. \textbf{Optimizing factuality}. We evaluate how our method performs when optimizing factuality rather than ROUGE or BLEU.  This is achieved by using a model that measures factuality ~\cite{factmodel} as the reward function. In this setting,  we directly use ground truth as the static dataset. Tab.~\ref{tab:fact} shows that our method achieves the best factuality score. For the ROUGE scores that are not directly optimized,  we still achieve the best or second-best performance.

2. \textbf{Optimizing click-through rate (CTR)}. We add an advertisement generation task ~\cite{umpg},  in which the goal is to generate a textual advertisement based on the textual description on the product website. The reward is the click-through rate given by a click prediction model. We also use ground-truth as the static dataset. Tab.~\ref{tab:ctr} shows that our method can achieve good results in a diverse set of metrics,  including ROUGE,  Distinct,  and CTR. Here,  Distinct measures the ratio of distinct uni-grams in an output.

\section{Additional Criteria including Human Evaluation Scores.} 
\label{appendix:human}
For SAMSum, we add other criteria of quality that have not been directly optimized,  including human evaluation score and BertScore ~\cite{bertscore} that measures the similarity to ground-truth in the latent space. In human evaluation,  we give the guidelines by following~\cite{gliwa-etal-2019-samsum}. In particular,  we ask workers to score the overall quality from -1,  0,  and 1 (e.g.,  1 stands for ``it is understandable and gives a brief overview of the text,''  while -1 means that ``a summarization is poor,  extracts irrelevant information or does not make sense at all''). We sample 50 instances in the test set,  ask 3 workers to score the outputs of different models for each instance,  and report the average score here. As shown in Tab.~\ref{tab:human_sam},  in addition to the ROUGE scores which we directly optimize,  we also perform the best in human evaluation and BertScore when compared with the most competitive baseline (AVG) and the base model. 
\begin{table}[h]
\vspace{-5mm}
\caption{More metrics and human evaluation on SAMSum. $*$ indicates the metrics directly Optimized during training. }
\label{tab:human_sam}
\vspace{-3mm}
\begin{center}
\begin{small}
\resizebox{1\linewidth}{!}{
\begin{tabular}{lccccc}
\toprule
 & Human & BertScore & R-1*& R-2* & R-L*\\ \midrule
Base  & 0.52             & 0.5637    & 53.09                    & 28.17                    & 49.02                    \\ 
AVG   & 0.57             & 0.5745    & 54.10                    & \textbf{29.21}                    & 49.58                    \\ 
OURS  & \textbf{0.63}    & \textbf{0.5762}    & \textbf{54.27}           & 29.19           & \textbf{50.57}           \\ 

\bottomrule
\end{tabular}
}
\end{small}
\end{center}
\vspace{-7mm}
\end{table}

\begin{table}[h]
\vspace{-3mm}
\caption{More metrics and human evaluation on SQuAD. $*$ indicates the metrics directly Optimized during training.}
\label{tab:human_squ}
\vspace{-4mm}
\begin{center}
\begin{small}
\resizebox{1\linewidth}{!}{
\begin{tabular}{lccccccc}
\toprule
&  Grammar & Understanding & Correctness  & B-4* & R-L* & MTR \\ \midrule
Base  & \textbf{2.91} &  2.95 & \textbf{2.71}   &  27.43                   & 54.30                     & 27.82                   \\ 
AVG   & 2.89 & 2.94  &  2.69  & 27.50                  & 54.79                   & 27.77                   \\ 
OURS  & \textbf{2.91} & \textbf{2.97}  & \textbf{2.71}  & \textbf{27.79}           & \textbf{54.95}           & \textbf{28.32}           \\ 

\bottomrule
\end{tabular}
}
\end{small}
\end{center}
\vskip -0.1in
\end{table}
For SQuAD, we follow ~\cite{ushio-etal-2022-generative} to include a humane evaluation based on  grammaticality, understandability and correctness with a 3-point scale.
In Tab.~\ref{tab:human_squ}, the human evaluation in SQuAD show a slight improvement. The reason may be that the base model can do this task well, as its score is already close to the perfect score of 3.



\end{document}